\documentclass[
 two-column
]{ceurart}

\usepackage{listings}
\usepackage{amsmath}
\usepackage{url}
\begin{document}

\copyrightyear{2022}
\copyrightclause{Copyright for this paper by its authors.
  Use permitted under Creative Commons License Attribution 4.0
  International (CC BY 4.0).}

\conference{Forum for Information Retrieval Evaluation, December 12-15, 2024, India}

\title{CryptoLLM: Unleashing the Power of Prompted LLMs for SmartQnA and Classification of Crypto Posts}

%\tnotetext[1]{You can use this document as the template for preparing your
%  publication. We recommend using the latest version of the ceurart style.}

\author[1]{Aniket Deroy}[%
orcid=0000-0001-7190-5040,
email=roydanik18@kgpian.iitkgp.ac.in,
%url=https://yamadharma.github.io/,
]
\cormark[1]
\fnmark[1]
\address[1]{IIT Kharagpur,
  Kharagpur, India}
%\address[2]{}%Joint Institute for Nuclear Research,
  %6 Joliot-Curie, Dubna, Moscow region, 141980, Russian Federation}

\author[1]{Subhankar Maity}[%
orcid=0009-0001-1358-9534,
email=subhankar.ai@kgpian.iitkgp.ac.in,
%url=https://kmitd.github.io/ilaria/,
]
%\address[3]{Vrije Universiteit Amsterdam, De Boelelaan 1105, 1081 HV Amsterdam, The Netherlands}

\cortext[1]{Corresponding author.}
%\fntext[1]{These authors contributed equally.}

\begin{abstract}
The rapid growth of social media has resulted in an large volume of user-generated content, particularly in niche domains such as cryptocurrency. This task focuses on developing robust classification models to accurately categorize cryptocurrency-related social media posts into predefined classes, including but not limited to objective, positive, negative, etc. Additionally, the task requires participants to identify the most relevant answers from a set of posts in response to specific questions. By leveraging advanced LLMs, this research aims to enhance the understanding and filtering of cryptocurrency discourse, thereby facilitating more informed decision-making in this volatile sector. We have used a prompt-based technique to solve the classification task for reddit posts and twitter posts. Also, we have used 64-shot technique along with prompts on GPT-4-Turbo model to determine whether a answer is relevant to a question or not.

\end{abstract}

\begin{keywords}
  GPT \sep
  Relevance \sep
  Classification \sep
  Few-shot \sep
  Prompt Engineering 
\end{keywords}


\maketitle

\section{Introduction}

We have seen the emergence of cryptocurrencies has generated substantial interest and discourse across various social media platforms \cite{cr1}. With millions of users expressing their opinions, sharing information, and speculating on the future of digital assets, the need for effective tools to process and analyze this data has become increasingly critical \cite{cr2}. However, the diverse nature of social media content, characterized by informal language, varying tones, and mixed sentiments, presents significant challenges for traditional text classification methods \cite{cr3}.

This task is designed to address these challenges by inviting participants to develop machine learning models capable of accurately classifying cryptocurrency-related social media posts into one of eight categories, such as objective, positive, negative, and others. Beyond simple classification, the task also extends to the identification of relevant answers to specific questions from a corpus of posts, a function that mirrors real-world applications where users seek precise information in a sea of data \cite{cr4}.

By focusing on the cryptocurrency domain, this task~\cite{cryptoqa2024} not only contributes to the broader field of sentiment analysis and text classification but also offers practical implications for stakeholders in the financial sector. Improved classification and information retrieval from social media can lead to better market insights, more effective communication strategies, and enhanced user experiences in digital finance. This study aims to showcase the potential of natural language processing techniques in transforming raw social media data into actionable intelligence in the context of cryptocurrency \cite{cr5}.
In our approach to solving the classification task for Reddit and Twitter posts, we employed a prompt-based technique \cite{cm25}, which has proven to be effective in leveraging the strengths of large language models for text classification. Specifically, we utilized carefully designed prompts to guide the model in understanding the context and nuances of social media posts related to cryptocurrency. This approach enabled us to harness the power of GPT-4-Turbo \cite{cr6} to classify posts into predefined categories, such as objective, positive, negative, etc.

To enhance the model's ability to determine the relevance of answers to specific questions, we implemented a 64-shot technique \cite{cr7} in conjunction with our prompt-based method to determine whether a answer is relevant to a question or not. %The 64-shot technique involved providing the model with 64 example pairs of questions and relevant or irrelevant answers during the fine-tuning phase.
This allowed the model to learn from a variety of examples, thereby improving its accuracy in identifying whether a given answer is pertinent to the corresponding question.

%By combining the prompt-based technique with the 64-shot method on the GPT-4 model, we were able to achieve a high level of performance in both the classification and relevance determination tasks. This approach not only improved the model's understanding of the context within social media posts but also enhanced its ability to distinguish between relevant and irrelevant information in response to specific queries.

%We observe that for Tamil and Kannada, GPT models have significant room for improvement.

\section{Related Work}

The task of classifying and analyzing social media posts, particularly in specialized domains such as cryptocurrency, has garnered significant attention in recent years \cite{cr3}. Social media platforms like Twitter and Reddit are rich sources of user-generated content that reflect public sentiment, market trends, and community discussions \cite{cr8}. However, the unstructured and often noisy nature of social media data presents challenges for traditional text classification and information retrieval methods \cite{cr9}. Early efforts in text classification on social media data primarily relied on traditional machine learning approaches, such as support vector machines (SVMs) and logistic regression, combined with handcrafted features like term frequency-inverse document frequency (TF-IDF) and n-grams \cite{cr11, cr10, cr12, cr13}. However, these approaches often struggled with the informal and context-dependent language prevalent on platforms like Twitter and Reddit.

With the advent of deep learning, particularly with the development of convolutional neural networks (CNNs) and recurrent neural networks (RNNs), the field saw a shift towards more robust models that could capture the nuances of language more effectively \cite{cr14, cr15, cr16, cr17, cr18}. More recently, transformer-based models like BERT (Bidirectional Encoder Representations from Transformers) have set new benchmarks in text classification by leveraging contextual embeddings to better understand the intricacies of language \cite{cr19, cr20}. These models have been particularly effective in sentiment analysis, where understanding the context and tone of a post is crucial.

%Prompt-Based Techniques:
Prompt-based learning \cite{cm25}, as popularized by models such as GPT-3 \cite{cr7} and GPT-4-Turbo \cite{cr6}, has emerged as a powerful paradigm in NLP. Unlike traditional supervised learning approaches, prompt-based methods guide the model's predictions by framing the task as a completion or question-answering problem. This technique has been shown to be highly effective, particularly when dealing with limited labeled data, as it allows the model to leverage its pre-existing knowledge more effectively . In our work, we build on this approach by designing specific prompts tailored to the cryptocurrency domain, enhancing the model's ability to classify posts and identify relevant information.

%Few-Shot Learning:
Few-shot learning techniques \cite{cr7} have gained traction as a means to improve model performance in scenarios where labeled data is scarce. In particular, the use of n-shot learning, where the model is exposed to a small number of labeled examples during training, has shown promise in various NLP tasks \cite{cr26}. The 64-shot technique we employed is an extension of this concept, allowing the model to learn from a diverse set of examples, thereby improving its generalization to unseen data . Recent studies have demonstrated the effectiveness of few-shot learning in improving the performance of large language models (LLMs) on various tasks, including text classification \cite{cr21, cr22, cr23} and question answering \cite{cr24, cr25}.

%Cryptocurrency-Related Social Media Analysis:
Research specifically targeting cryptocurrency-related social media content has been limited but is growing in importance as digital currencies become more mainstream. Previous studies have primarily focused on sentiment analysis \cite{cr14, cr15}, attempting to correlate social media sentiment with cryptocurrency price movements . However, there has been a lack of comprehensive studies that address both classification and relevance determination in this domain. Our work contributes to filling this gap by combining prompt-based learning and few-shot techniques to handle the unique challenges presented by cryptocurrency-related social media posts. In summary, our work builds on the advancements in LLMs, prompt engineering, and few-shot learning to address the specific challenges of classifying and retrieving relevant information from cryptocurrency-related social media data. By integrating these techniques, we aim to push the boundaries of what can be achieved in this specialized and dynamic field.

\section{Dataset}

%The QnA task has 5058 posts in the train set. The Twitter opinion dataset has 17797 posts in the train set. The Reddit opinion task has 25431 posts in the train set.

The QnA task has 6323 posts in the test set. The Twitter opinion dataset has 500 posts in the test set. The Reddit opinion dataset has 500 posts in the test set.

We use the reddit opinion dataset in Zero-shot setting.
We use the twitter opinion dataset in Zero-shot setting.
We use the QnA dataset in 64-shot setting.
%\end{enumerate}
Since we primarily use only the test set data for our predictions we only mention the statistics of the test set data.

\section{Task Definition}
%Objective:
The Task-1 is to develop a classification model to classify cryptocurrency related social media posts into eight classes namely, \textit{Noise}, \textit{Objective}, \textit{Positive}, \textit{Negative}, \textit{Neutral}, \textit{Question}, \textit{Advertisement}, \textit{Miscellaneous}.

The Task-2 is, for a given question wrt cryptocurrency related post we have to detect all the answers which are relevant to this question.

\section{Methodology}

\subsection{Why Prompting?}

\begin{itemize}[-]
   
\item \textbf{Leveraging Pretrained Language Model Knowledge:}
Prompting allows us to directly utilize the extensive knowledge embedded in large pre-trained language models like GPT-4-Turbo \cite{cm25}. These models are trained on vast and diverse datasets, capturing a wide range of linguistic patterns, contextual nuances, and domain-specific information. By crafting appropriate prompts, we can guide the model to apply this knowledge effectively to specific tasks, such as classifying social media posts or finding relevant answers \cite{cm25}.

\item \textbf{Reducing the Need for Large Labeled Datasets:}
Traditional machine learning approaches often require substantial labeled datasets to achieve high performance \cite{cr27}. However, obtaining large, high-quality labeled datasets, especially in specialized domains like cryptocurrency, can be resource intensive. Prompting, especially when combined with few-shot learning techniques, mitigates this need by enabling the model to perform well with minimal labeled examples \cite{cr28}. This is particularly beneficial in scenarios where labeled data is scarce or costly to acquire.

\item \textbf{Flexibility and Adaptability:}
Prompting offers significant flexibility in task formulation. By simply changing the structure or wording of a prompt, we can adapt the model to perform a wide variety of tasks without the need for retraining \cite{cm25}. This adaptability is crucial when dealing with dynamic content, such as social media posts, where topics, language, and context can rapidly evolve.

\item \textbf{Improved Contextual Understanding:}
Social media posts, especially those related to cryptocurrency, often contain informal language, abbreviations, slang, and domain-specific jargon \cite{cr29}. Traditional classification models might struggle with this variability. Prompt-based approaches, however, leverage the model's deep contextual understanding, allowing it to better interpret and classify such content \cite{cm25}. For instance, a well-crafted prompt can help the model distinguish between positive and negative sentiment even when the language is complex or non-standard.

\item \textbf{Enhanced Relevance Determination:}
Finding relevant answers to specific questions within a corpus of social media posts is a complex task, as it requires understanding the subtle relationships between questions and potential answers \cite{cr30}. Prompting allows us to frame this task in a way that aligns with the model's strengths, treating it as a natural language inference problem where the model evaluates the likelihood that a given answer is relevant to the question \cite{cm25}. Using prompt-based few-shot learning, we can further fine-tune this ability, improving the accuracy of relevance determination in contexts where direct answer-question pairs are not always straightforward \cite{cr25}.

\item \textbf{Rapid Prototyping and Experimentation:}
Prompt-based techniques facilitate rapid prototyping and experimentation, allowing researchers and practitioners to quickly test different approaches to a problem \cite{cr31}. This is particularly useful in the fast-paced world of cryptocurrency, where market conditions and public sentiment can change rapidly \cite{cr32}. Prompting enables quick adjustments to the behavior of the model, ensuring that it remains effective even as the underlying data evolves \cite{cm25}.

\item \textbf{Alignment with Human-Like Reasoning:}
Prompt-based methods are designed to mimic how humans might approach a task, framing it naturally and intuitively \cite{cr33}. This alignment with human-like reasoning makes the model's predictions more interpretable and reliable, particularly in tasks like classification and relevance determination where understanding subtle nuances is crucial \cite{cr34}. By asking the model to "complete" or "answer" in a way that mirrors human thought processes, we can achieve more accurate and contextually appropriate outcomes.

\end{itemize}
In summary, prompting offers a powerful and efficient way to solve classification tasks and find relevant answers in complex, dynamic, and data-scarce environments like cryptocurrency-related social media. It capitalizes on the strengths of large language models, reducing the need for extensive labeled data, and providing flexibility, adaptability, and improved performance across a wide range of tasks.

\subsection{Prompt Engineering along with Few-shot Methods}

We used the GPT-4-Turbo model via prompting through the OpenAI API\footnote{\url{https://platform.openai.com/docs/models/gpt-4-turbo-and-gpt-4}} to solve the classification task. After the prompt is provided to the LLM, the following steps happen internally while generating the output. We outline the steps that occur internally within the LLM (i.e., a summary of the prompting approach using GPT-4 Turbo).\\

\textbf{Step 1: Tokenization}

\begin{itemize}
    \item \textbf{Prompt:} \( X = [x_1, x_2, \dots, x_n] \)
    \item The input text (prompt) is first tokenized into smaller units called tokens. These tokens are often subwords or characters, depending on the model's design.
    \item \textbf{Tokenized Input:} \( T = [t_1, t_2, \dots, t_m] \)
\end{itemize}

\textbf{Step 2: Embedding}

\begin{itemize}
    \item Each token is converted into a high-dimensional vector (embedding) using an embedding matrix \( E \).
    \item \textbf{Embedding Matrix:} \( E \in \mathbb{R}^{|V| \times d} \), where \( |V| \) is the size of the vocabulary and \( d \) is the embedding dimension.
    \item \textbf{Embedded Tokens:} \( T_{\text{emb}} = [E(t_1), E(t_2), \dots, E(t_m)] \)
\end{itemize}

\textbf{Step 3: Positional Encoding}

\begin{itemize}
    \item Since the model processes sequences, it adds positional information to the embeddings to capture the order of tokens.
    \item \textbf{Positional Encoding:} \( P(t_i) \)
    \item \textbf{Input to the Model:} \( Z = T_{\text{emb}} + P \)
\end{itemize}

\textbf{Step 4: Attention Mechanism (Transformer Architecture)}

\begin{itemize}
    \item \textbf{Attention Score Calculation:} The model computes attention scores to determine the importance of each token relative to others in the sequence.
    \item \textbf{Attention Formula:}
    \begin{equation}
    \text{Attention}(Q, K, V) = \text{softmax}\left(\frac{QK^T}{\sqrt{d_k}}\right)V
    \end{equation}
    \item where \( Q \) (query), \( K \) (key), and \( V \) (value) are linear transformations of the input \( Z \).
    \item This attention mechanism is applied multiple times through multi-head attention, allowing the model to focus on different parts of the sequence simultaneously.
\end{itemize}

\textbf{Step 5: Feedforward Neural Networks}

\begin{itemize}
    \item The output of the attention mechanism is passed through feedforward neural networks, which apply non-linear transformations.
    \item \textbf{Feedforward Layer:}
    \begin{equation}
    \text{FFN}(x) = \max(0, xW_1 + b_1)W_2 + b_2
    \end{equation}
    \item where \( W_1, W_2 \) are weight matrices and \( b_1, b_2 \) are biases.
\end{itemize}

\textbf{Step 6: Stacking Layers}

\begin{itemize}
    \item Multiple layers of attention and feedforward networks are stacked, each with its own set of parameters. This forms the "deep" in deep learning.
    \item \textbf{Layer Output:}
    \begin{equation}
    H^{(l)} = \text{LayerNorm}(Z^{(l)} + \text{Attention}(Q^{(l)}, K^{(l)}, V^{(l)}))
    \end{equation}
    \begin{equation}
    Z^{(l+1)} = \text{LayerNorm}(H^{(l)} + \text{FFN}(H^{(l)}))
    \end{equation}
\end{itemize}

\textbf{Step 7: Output Generation}

\begin{itemize}
    \item The final output of the stacked layers is a sequence of vectors.
    \item These vectors are projected back into the token space using a softmax layer to predict the next token or word in the sequence.
    \item \textbf{Softmax Function:}
    \begin{equation}
    P(y_i|X) = \frac{\exp(Z_i)}{\sum_{j=1}^{|V|} \exp(Z_j)}
    \end{equation}
    \item where \( Z_i \) is the logit corresponding to token \( i \) in the vocabulary.
    \item The model generates the next token in the sequence based on the probability distribution, and the process repeats until the end of the output sequence is reached.
\end{itemize}

\textbf{Step 8: Decoding}

\begin{itemize}
    \item The predicted tokens are then decoded back into text, forming the final output.
    \item \textbf{Output Text:} \( Y = [y_1, y_2, \dots, y_k] \)
\end{itemize}

%Next we will discuss the prompts which are provided to the LLMs:

\begin{figure}[h!]
  \centering
  \includegraphics[width=0.50\linewidth]{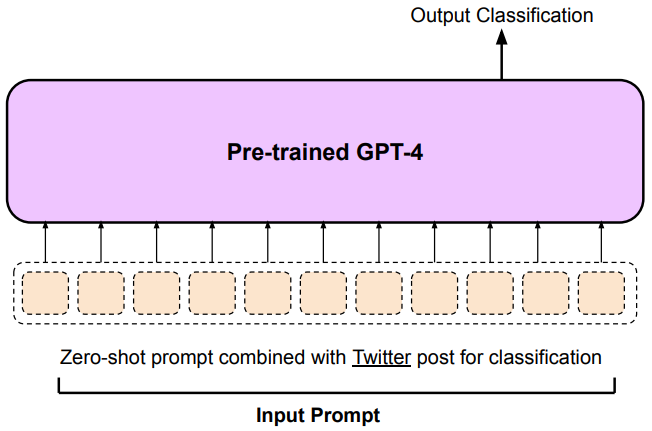}
  \caption{An overview of the methodology for classifying Twitter posts using GPT-4-Turbo.} \label{fig1}
  %\Description{A woman and a girl in white dresses sit in an open car.}
\end{figure}

The prompt for classifying Twitter posts: \textit{"Classify the Twitter post <Twitter\_post> into one of the following labels: Noise, Objective, Negative, Positive, Neutral, Question, Advertisement, Miscellaneous"}. The overview diagram for the classification of Twitter posts is presented in Figure ~\ref{fig1}.

\begin{figure}[h!]
  \centering
  \includegraphics[width=0.50\linewidth]{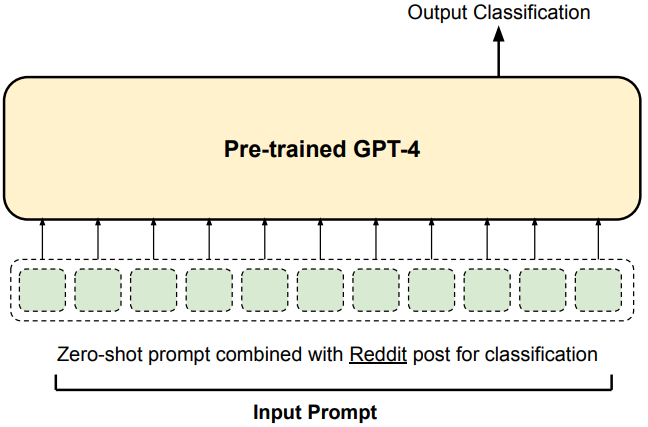}
  \caption{An overview of the methodology for classifying Reddit posts using GPT-4-Turbo.} \label{fig2}
  %\Description{A woman and a girl in white dresses sit in an open car.}
\end{figure}

The prompt for classifying Reddit posts: \textit{"Classify the Reddit post <Reddit\_post> into one of the following labels: Noise, Objective, Negative, Positive, Neutral, Question, Advertisement, Miscellaneous"}. The overview diagram for the classification of Reddit posts is depicted in Figure ~\ref{fig2}.
%\hline

\begin{figure}[h!]
  \centering
  \includegraphics[width=0.50\linewidth]{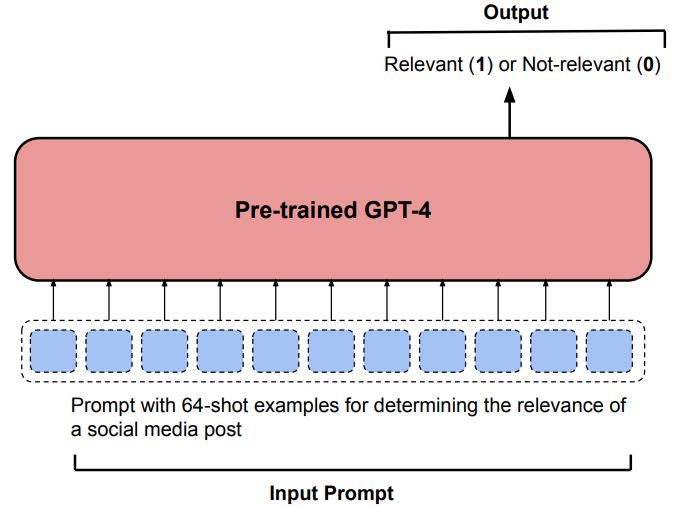}
  \caption{An overview of the methodology for determining the relevance of a social media post using GPT-4-Turbo.} \label{fig3}
  %\Description{A woman and a girl in white dresses sit in an open car.}
\end{figure}

Along with the 64-shot example, the following prompt will be provided, \textit{"Given the title <title> and comment <comment\_body>. Please check whether the comment is relevant or not-relevant to the title. Only state relevant or not-relevant."} The overview diagram for determining relevance of posts is shown in Figure ~\ref{fig3}.

For the results reported, we ran the GPT-4-Turbo model at a temperature of 0.7.
%\hline

%After getting the relevance score we used the following mathematical formulation to allow for sequential presence of relevant documents,

%if the score of the current document %\(D_n\) is less than 0.1, and you want the probability for the next document to be just the score itself, the formulation can be updated with a conditional expression.
%This can be written as follows:

%\[
%P(D_{n+1} \mid D_n) =
%\begin{cases} 
%\text{Score}(D_{n+1}) & \text{if } %\text{Score}(D_{n+1}) < 0.1 \\
%\text{Score}(D_{n+1}) & \text{if } n = -1 \\
%0.3 + \text{Score}(D_{n+1}) & \text{if } D_n = relevant 
%\end{cases}
%\]

%This equation now reflects that if the score of the current document \(D_n\) is less than 0.1, the probability of the current document being relevant is simply equal to the relevance score of current document.
%If the previous document is relevant then the probability that the current document is relevant is (0.3 + Score) for the current document.
%For the first document, the probability is equal to the relevance score.

%If the probability score of a particular document is greater than 0.6 we consider the document to be relevant to the query. Like this we found out all documents which are relevant to a query.

%\(\text{Score

%The diagram for GPT-3.5 Turbo is shown in Figure ~\ref{fig1}.

%\noindent \textbf{(a)} We used the following prompt for Kannada language for the purpose of classification: 

%"\textit{Please identify which category the word is in English, Kannada, Mixed, Name, Location, Symbol and Other. Please state En, Kn, Mixed, Name, Location, sym and Other. The word is <Word>}."

%The figure representing the methodology is shown in Figure ~\ref{fig2}.

%The figure representing the methodology is shown in Figure ~\ref{fig3}.

%For the results reported, we ran the GPT model at temperature value of 0.7. %ranging from: {0.5, 0,6, 0.7, 0.8, and 0.9}.

%\noindent \textbf{(b)} We used the following prompt for Tamil language for the purpose of classification: 

%"\textit{Please identify which category the word is in English, Tamil, Mixed, Name, Location, Symbol and Other. Please state en, tm, tmen, name, Location, sym and Other. The word is <Word>.}"

%The figure representing the methodology is shown in Figure ~\ref{fig3}.
\section{Results}
For our team named \textbf{TextTitans}, the Macro-F1 score for reddit opinion task is 0.249 with a rank of 8th. A score of 0.249 indicates that the model's performance is relatively low, possibly due to challenges in handling the specific data or class imbalances in this task.

The macro-F1 score for twitter opinion task is 0.266 with a rank of 8th.  The slightly higher score compared to the Reddit task suggests that the model performed marginally better on Twitter data but still struggled overall.

The macro-F1 for QnA task is 0.157 with a rank of 1st. For the QnA task, the model achieved a Macro-F1 score of 0.157, which is the lowest score among the three tasks. However, the model ranked 1st in this task, indicating that, despite the low score, it outperformed other models in this specific challenge. This could suggest that the QnA task is particularly difficult, leading to lower scores overall, but the model was still the best among its competitors.

\section{Conclusion}
%Conclusion
This study evaluated the performance of our model across three distinct tasks: Reddit opinion classification, Twitter opinion classification, and Question-and-Answer (QnA) task. The Macro-F1 scores for the Reddit and Twitter opinion tasks were 0.249 and 0.266, respectively, placing the model 8th in both tasks. These results highlight the challenges our model faced in effectively handling the varied and complex nature of social media data, possibly due to class imbalances or the nuanced expressions of opinions on these platforms.

In contrast, the model achieved a Macro-F1 score of 0.157 in the QnA task, securing the 1st position. Despite the lower absolute score, this top ranking indicates that our model outperformed others in this particularly challenging task, suggesting its robustness in scenarios where the data complexity is high and the overall performance across models is lower.

These findings underscore the importance of tailoring models to the specific characteristics of each task and dataset. Future work will focus on refining the model to better capture the intricacies of social media language and enhancing its ability to handle a broader range of tasks with more balanced performance across all metrics.

%\begin{acknowledgments}
%  Thanks to the developers of ACM consolidated LaTeX styles
%  \url{https://github.com/borisveytsman/acmart} and to the developers
%  of Elsevier updated \LaTeX{} templates
%  \url{https://www.ctan.org/tex-archive/macros/latex/contrib/els-cas-templates}.  
%\end{acknowledgments}

%%
%% Define the bibliography file to be used
\bibliography{sample-ceur}

\appendix


%The sources for the ceur-art style are available via
%\begin{itemize}
%\item \href{https://github.com/yamadharma/ceurart}{GitHub},
% \item \href{https://www.overleaf.com/project/5e76702c4acae70001d3bc87}{Overleaf},
%\item
%  \href{https://www.overleaf.com/latex/templates/template-for-submissions-to-ceur-workshop-proceedings-ceur-ws-dot-org/pkfscdkgkhcq}{Overleaf
%    template}.
%\end{itemize}

\end{document}